\documentclass[conference]{IEEEtran} 
\usepackage[colorlinks=true,allcolors=blue]{hyperref}
\usepackage{amsmath,amssymb,xspace}
\usepackage{amsfonts}
\usepackage[T1]{fontenc}
\usepackage{algorithm}
\usepackage{algorithmic}
\usepackage{caption}
\usepackage{subcaption}
\usepackage{multirow,xspace}
\usepackage{tikz}
\usepackage{pgfplots}
\usepackage{wrapfig}
\usepackage[inline]{enumitem}
\usepackage{graphicx,color,stackrel,xcolor}
\usepackage{color}

\definecolor{darkred}{RGB}{159,67,67}
\definecolor{lightgreen}{RGB}{100,255,100}
\usepackage{pgf,tikz}
\usetikzlibrary{shapes,fit}
\usetikzlibrary{arrows,matrix}
\usetikzlibrary{positioning}
\usetikzlibrary{arrows,automata,calc,positioning}
\usetikzlibrary{shapes.arrows,decorations.pathreplacing,backgrounds}
\usetikzlibrary{patterns}
\tikzstyle{component}=[draw, text centered, text width=5em]
\tikzstyle{arbitrer}=[draw, circle, text centered, text width=1em]
\tikzstyle{ann} = [above, text width=5em]

\newcommand{\lard}{{\sc lard}\xspace}

\newcommand{\GES}{Google Earth Studio\xspace}
\newcommand{\flightsim}{Microsoft Flight Simulator\xspace}

\definecolor{turquoise}{rgb}{.45,.93,.90}
\definecolor{redish}{rgb}{.95,.40,.45}
\definecolor{greenish}{rgb}{.63,.89,.60}

\newtheorem{definition}{\textbf{Definition}}
\newtheorem{mytask}{\textbf{Task}}
\newtheorem{intended}{\textbf{Intended function}}
\newtheorem{odd}{\textbf{Operational Design Domain}}
\newtheorem{dqr}{\textbf{DQR}}

\begin{document}
\title{How to design a dataset compliant with an ML-based system ODD?}
\author{
    \IEEEauthorblockN{
        Cyril Cappi\IEEEauthorrefmark{1},
        Noémie Cohen\IEEEauthorrefmark{2},
        M\'elanie Ducoffe\IEEEauthorrefmark{2},
        Christophe Gabreau\IEEEauthorrefmark{2},
        Laurent Gardes\IEEEauthorrefmark{1},
        Adrien Gauffriau\IEEEauthorrefmark{2},\\
        Jean-Brice Ginestet\IEEEauthorrefmark{3},
        Franck Mamalet\IEEEauthorrefmark{4},
        Vincent Mussot\IEEEauthorrefmark{4},
        Claire Pagetti\IEEEauthorrefmark{5},
        David Vigouroux\IEEEauthorrefmark{4}
    }
    \IEEEauthorblockA{\IEEEauthorrefmark{1}SNCF, \IEEEauthorrefmark{2}Airbus, \IEEEauthorrefmark{3}DGA, \IEEEauthorrefmark{4}IRT Saint Exupéry, \IEEEauthorrefmark{5}ONERA}
}

\maketitle
\begin{abstract}
This paper focuses on a Vision-based Landing task and presents the design and the validation of a dataset that would comply with the Operational Design Domain (ODD) of a Machine-Learning (ML) system. 
Relying on emerging certification standards, we describe the process for establishing ODDs at both the system and image levels. In the process, we present the translation of high-level system constraints into actionable image-level properties, allowing for the definition of verifiable Data Quality Requirements (DQRs).
To illustrate this approach, we use the Landing Approach Runway Detection (\lard) dataset which combines synthetic imagery and real footage, and we focus on the steps required to verify the DQRs.
The replicable framework presented in this paper addresses the challenges of designing a dataset compliant with the stringent needs of ML-based systems certification in safety-critical applications.

\end{abstract}
\section{Introduction}

Artificial Intelligence (AI) is rapidly becoming a cornerstone technology in various sectors, including transportation. In aeronautics, AI promises efficiency enhancement and operational cost reductions, yet its adoption remains complex. This is primarily due to the stringent certification process these systems must undergo to meet the rigorous safety standards of the domain. Thus, this paper delves into the challenges of certifying AI in aviation, focusing on the design of a dataset that would comply with the Operational Design Domain (ODD) of an AI-based system.


\subsection{Certification guidelines}
EUROCAE WG-114/SAE G-34 is a joint working group 
on the certification of ML-based systems 
that will release the ED-324/ARP-6983 soon. 
Even if not yet published, 
there are several publications
\cite{gabreau:hal-WG,gabreau:hal-03761946,kaakai2023datacentric} that highlight the objectives and activities expected by the Aerospace Recommended Practice (ARP).
In parallel, the European certification authorities --
EASA -- published their concept papers
\cite{Easaconcept,Easaconcept2}
that aim at guiding applicants introducing AI (Artificial Intelligence) / ML (machine learning) technologies into systems intended for use in safety-related or environment-related applications.

Both guidelines rely on the existing standards as much as possible. From an airborne perspective, this means using the ED-79/ARP-4754A \cite{arp4754} guidance whenever possible to integrate the ML-based function at subsystem level and using the ED-12C/DO-178C \cite{do178c} and the ED-80/DO-254 \cite{do254} when it comes to the deployment of the ML models onto respectively software and hardware items. The change of paradigm that comes with a data-driven development method entails a new process that covers the whole spectrum of ML-based system development.

In this work, we only focus on part of the new process called \emph{data management} to produce a dataset whose internal features match the intended function and its operating environment. 
Practically, the intended function must be defined with its Operational Design Domain (ODD). Then, the question is how to design a dataset compliant with such an ODD.

\subsection{Motivation}
To support our work, we have selected a safety-critical application, namely visual-based landing (VBL). 
Indeed, increasing the level of autonomy of aircraft will ease the flying in case of pilots' cognitive load and would therefore improve the safety in civil aviation. 
In a future where it is envisaged to fly with only one pilot on board, a single pilot may not be in capacity to assume all tasks required during the landing phase (especially the final ones). Thus, vision-based landing systems could circumvent such a need 
and would be in charge of computing the position of the aircraft from the position of the runway within an image taken 
during the landing phase of an aircraft. 
We particularly focus on the sub-task that consists of detecting the runway in the image.

\begin{figure}[hbt]
    \centering
    \begin{subfigure}[t]{.32\linewidth}
        \centering
        \includegraphics[width=\textwidth]{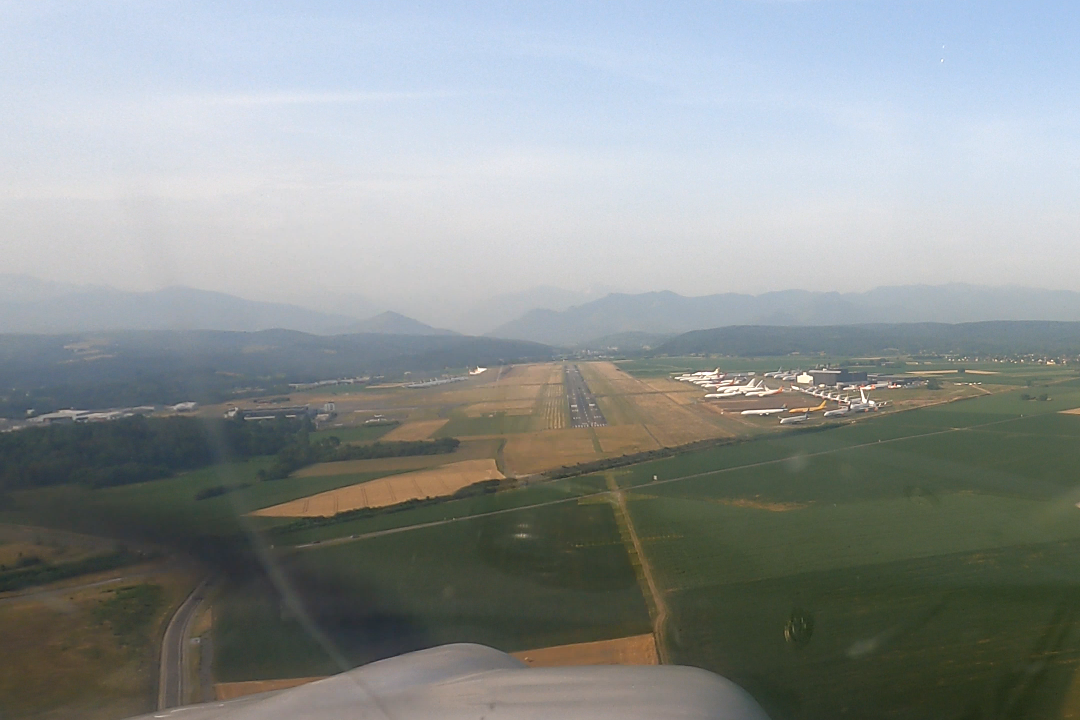}
    \end{subfigure}
    \hfill
    \begin{subfigure}[t]{.32\linewidth}
        \centering
        \includegraphics[width=\textwidth]{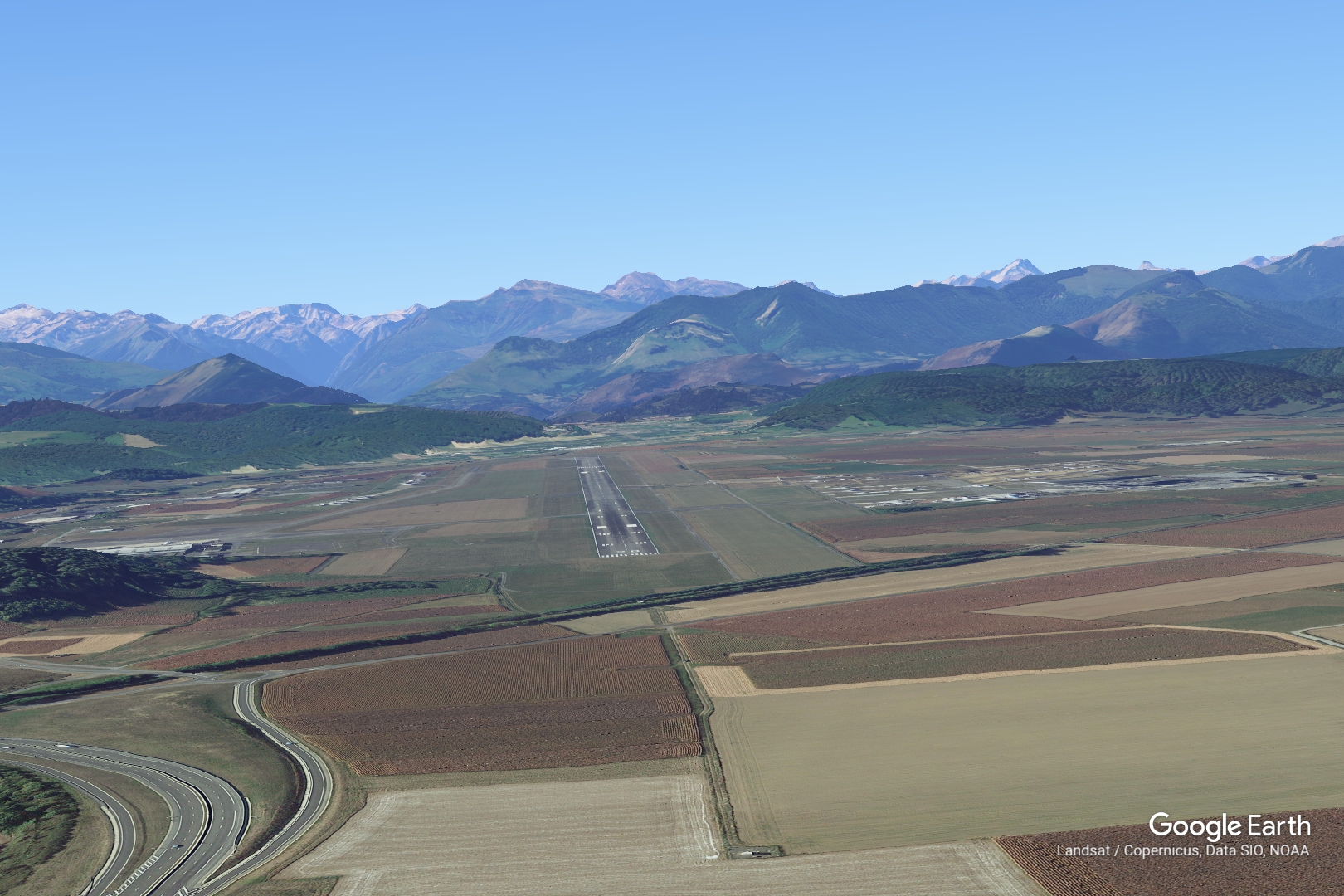}
    \end{subfigure}
    \hfill
    \begin{subfigure}[t]{.32\linewidth}
        \centering
        \includegraphics[width=\textwidth]{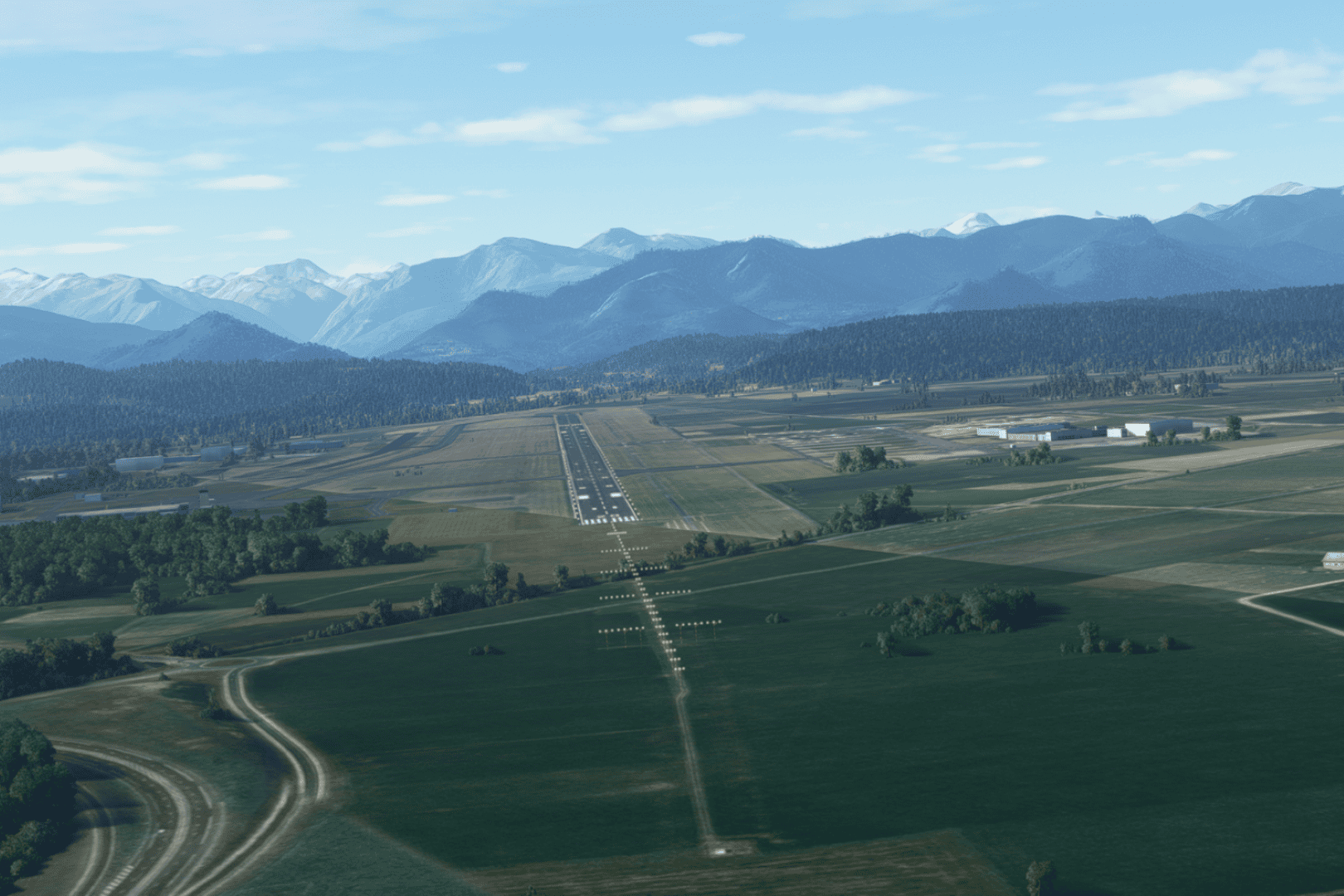}
    \end{subfigure}
    \caption{\small Illustration of the quality of the synthetic images - {Comparison of a real landing footage (left) with  synthetic replicas (\GES center, \flightsim right)}}
    \label{fig:real_vs_synt_tarbes}
\end{figure}

As no open-source use cases (and no dataset) were available at the beginning of our work, we first had to define what a visual-based landing system should be and how machine learning could help. 
This has lead us to develop the Landing Approach Runway Detection (\lard) open-source dataset \cite{ducoffe:hal-04056760}\footnote{\url{https://github.com/deel-ai/LARD}}. 
Among the important features of this dataset, we can cite the capacity to generate easily new data thanks first to synthetic data generators based on \GES and \flightsim, and second the selection of Youtube channels (such as \footnote{\url{https://www.youtube.com/user/TheGreatFlyer}})
with real landing footage video from which we can label new data easily. 
The Figure~\ref{fig:real_vs_synt_tarbes} reproduces an image recorded during a flight and with our generators. Although the weather conditions differ between the images, we note a great similarity in the runway's environment.

\subsection{Approach}
\begin{figure}[hbt]
    \centering
    \resizebox{.7\linewidth}{!}{\begin{tikzpicture}[thick,scale=0.5]
\tikzstyle{block} = [draw,minimum height=2em, minimum width=1.5cm, inner sep=3pt];
\tikzstyle{txt} = [text centered, inner sep=0pt];

\path (0,-2) node[block] (syst_odd) {
\begin{tabular}{c} system-level\\ ODD \end{tabular}
};

\path (syst_odd) + (8,0) node (conops) [rounded corners, draw, thick, inner sep=2pt] {\begin{tabular}{l} \textbf{CONOPS}\\ - operating conditions\\- VBL: Landing cone \\ - operational scenarios\end{tabular}};
      
\draw[<-, dashed, ultra thick] (syst_odd) -- (conops);

\path (syst_odd) + (0,-4) node[block] (MLC) {VBL constituent};
\path (MLC) + (8,0) node (archi) [rounded corners, draw, thick, inner sep=2pt] {\begin{tabular}{l} - intended function\\- architecture\end{tabular}};

\draw[<-, dashed, ultra thick] (MLC) -- (archi);

\path (MLC) + (0,-4) node[block] (mlcodd) {
\begin{tabular}{c} MLCOOD\\ Image-level\\ODD \end{tabular}
};

\path (mlcodd) + (8,0) node (im_odd) [rounded corners, draw, thick, inner sep=2pt] {\begin{tabular}{l} - projection of system-\\ \hspace{0.3cm}level ODD on image\\ \hspace{0.3cm}via MLC architecture\\- experts concepts\\- DQRs \end{tabular}};

\draw[<-, dashed, ultra thick] (mlcodd) -- (im_odd);

\path (mlcodd) + (0,-5.5) node[block] (dataset) {
\begin{tabular}{c} VBL dataset\\ design and\\verification \end{tabular}
};

\path (dataset) + (8,0) node (lard) [rounded corners, draw, thick, inner sep=2pt] {\begin{tabular}{l} \textbf{LARD}\\ - projection of image\\ \hspace{0.3cm}level ODD\\- synthetic images\\ - real-footage\\ \hspace{0.3cm}of landing \end{tabular}};

\draw[<-, dashed, ultra thick] (dataset) -- (lard);

\draw[->, ultra thick] (syst_odd) -- (MLC);
\draw[->, ultra thick] (MLC) -- (mlcodd);
\draw[->, ultra thick] (mlcodd) -- (dataset);
\end{tikzpicture}}
    \caption{ODD Design Workflow}
    \label{fig:ODD_workflow}
\end{figure}
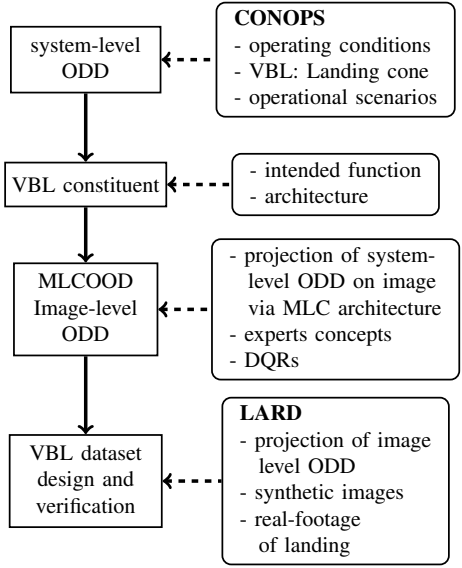

We propose a preliminary approach to apply the aeronautical certification guidelines. 
We have drawn an overview of our interpretation of the
ED-324/ARP6983 workflow, see figure \ref{fig:ODD_workflow}, to design a dataset from a system-level ODD provided by the ED-79/ARP-4754 and the system CONOPS.
 For the VBL system, the system-level ODD, presented in section \ref{sec-syst_odd}, consists of the landing approach geometry of an aircraft and the standardised runway markings. 
The ED-324/ARP6983 introduces an intermediate level of engineering (called ML Constituent -- MLC) between the system and item layers. From a system perspective, the MLC is an item (or a container of items). 
The MLC contains at least one ML model and 
its implementation should provide the necessary items to support the ML model(s) inference. 
The VBL constituent, presented in section \ref{sec-MLC},
contains three stages, among which an object detection stage for which we design the dataset.

The guidance advocates a refinement of the ODD that starts at the system level with the definition of the operating environment of the VBL system (landing approach conditions) and continues at the MLC level (MLCODD or image-level ODD in this use case) with some expected properties on the ML tasks. 
This activity is highly complex since we have to project the geometry of a landing on the possible images observed by the camera. 
We propose 3 activities, see section \ref{sec-image-level-odd}, to define the image-level ODD:
1) the geometric projection of a landing on the image;
2) the definition with the help of domain experts of \emph{expert concepts}; 
and finally 3) the definition of \emph{Data Quality Requirements} 
(DQRs) associated to the VBL. These DQRs are imposed by the ED-324/ARP6983 as a result of the ODD characterization activity.
Based on this image-level ODD, 
we developed a strategy for constructing the dataset to encompass those DQRs and we evaluated 
the compliance of the \lard dataset on some of them.
The \lard dataset definition and evaluation 
are detailed in section \ref{sec:dataset}.
Note that the workflow is highly iterative. Indeed, depending of the ability to refine the ODD into image-level ODD or the capacity of the dataset to comply with the requirements, it may be necessary to revise the system-level ODD or the ML constituent architecture.

\section{System-level ODD}
\label{sec-syst_odd}
The system-level ODD regroups the requirements that must flown from the
ED-79/ARP-4754 and the specific \emph{Concept of Operations} (CONOPS) of the system under development down to the dataset and model designs.

\subsection{ODD concept}
The concept of Operational Design Domain (ODD) originated in the automotive industry as a way to define the specific operating conditions under which automated driving systems are designed to function. The ODD concept was first introduced in the SAE (Society of Automotive Engineers) J3016 standard \cite{j3016} to define levels of driving automation for on-road motor vehicles. Their definition of ODD was \emph{"the operating conditions under which a given driving automation system or feature thereof is specifically designed to function, including, but not limited to, environmental, geographical, and time-of-day restrictions, and/or the requisite presence or absence of certain traffic or roadway characteristics"}.  
The ISO-21448/SOTIF (Safety Of The Intended Functionality) standard \cite{sotif} 
focuses on the safety considerations within automotive autonomous vehicles and directly integrates the concept of ODD from the SAE J3061 standard.

A more recent automotive standard, the ISO 34503 \cite{iso34503} proposes some concepts and requirements to enable the definition of an ODD for an automated driving system.
The document, in particular, distinguishes the ODD
and the \emph{Target Operational Domain} (TOD) that refers to the real-world conditions that a system may encounter. As the TOD is not \emph{specifiable}, 
it can be seen as a superset of the ODD. It is up to the system design to specify the \emph{optimal ODD} to be as close as possible to the TOD,
but it is also their responsibility to ensure that the system is used on the ODD solely and deactivated otherwise. 
The standard also promotes the definition of \emph{operational scenarios} on which the safety assessment should rely to evaluate the final system.

For aeronautical applications, the European Union Aviation Safety Agency (EASA) has adapted the ODD concept from SAE J3016.  The ODD is defined in EASA Artificial Intelligence Concept Paper Issue 2 \cite{Easaconcept2} as \emph{"the operating conditions under which a given AI/ML constituent is specifically designed to function as intended, including but not limited to environmental, geographical, and/or time-of-day restrictions"}. While the EASA definition is similar to the SAE J3016 definition, it differs in that it applies specifically to AI/ML constituents within a larger system, rather than to the system as a whole. This reflects the importance of defining constraints and requirements on the data used during the learning process, implementation, and inference in operations for AI/ML constituents. 
Nonetheless, the ODD concept remains an important tool for ensuring the safety and reliability of automated systems in both aeronautical and automotive applications.

\subsection{ED-324/ARP6983 -- operating environment}
The data management requires an upstream process at the system level of engineering to define the operating environment of the VBL system. This definition is developed from the expertise of \emph{subject matter experts} (SMEs) who have a deep knowledge of the Concept of Operations (CONOPS) and who can define the operational envelope of the system, i.e. the system operating conditions and environment where the system is supposed to operate correctly.
An accurate definition of the operating environment is a prerequisite for AI scientists to start the data management process and to define the MLCODD (image-level ODD).

 \subsection{VBL system-level ODD}
Defining such an ODD is highly complex \cite{Koopman2019HowMO} and of vital importance. 
For a vision-based system, it details in particular the environmental and weather conditions (e.g. temperature, wind, visibility, precipitation, types of sensor noise); operational terrain (e.g., runway slope, runway roughness); operational infrastructure (e.g. fixed obstacles) and many other information. Such a list could be infinite depending on the level of details. Making this problem tractable in practice is generally accomplished by constraining the operational environment to a subset of all possible situations that could be dealt with by a human.

\begin{figure}[hbt]
    \centering
    \includegraphics[width =.7\linewidth]{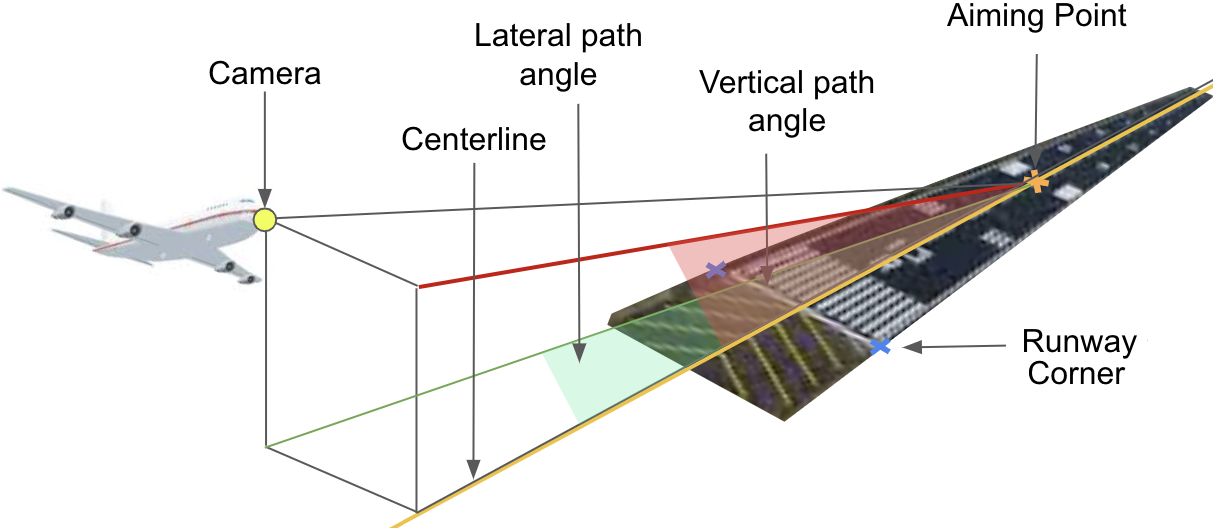}
    \caption{Geometry of a landing}
    \label{fig:camera_setting}
\end{figure}

The application is designed for civil aircraft landings.
Therefore, we start by defining a \emph{generic landing approach cone} based on the documentation provided by aeronautical standards. 
Figure~\ref{fig:camera_setting} illustrates 
the different positions / angles / distances / markings involved in the geometric description of a landing. Runway markings are standardized~\cite{runway_marking} and appear in most cases as follows: 
a first line at the start of the runway, called \emph{landing threshold}, represents the underline limit of the runway. It is usually followed by a pattern of stripes (the \emph{piano}) and then the runway identifiers. The target of an aircraft during landing is the \emph{Aiming Point}, located 300 meters beyond the landing threshold, between two rectangular markings visible on each side of the runway \emph{centerline}\footnote{An imaginary line going through the middle of the runway}.

The position of the aircraft with respect to the runway is defined by 3 parameters:
the along-track distance which corresponds to the distance between the projection of the aircraft nose on the centerline of the runway (on the ground) and the Aiming Point.
The lateral (resp. vertical) path angle which corresponds to the angle formed by the centerline and the line defined by the Aiming Point and the plane nose projection on the ground (resp. plane orthogonal to the ground going through the centerline).
On the other hand, the attitude of the aircraft is defined by its rotation angles 
(denoted respectively as pitch, roll, yaw). The yaw angle is relative to the runway heading\footnote{For instance a yaw of 0° indicates that the aircraft faces directly the runway, regardless of the runway orientation.} whereas pitch and roll are relative to the horizontal plane.

\begin{table}[t]
    \centering    
    \resizebox{.6\columnwidth}{!}{
        \begin{tabular}{|l|l|}
            \hline
            Parameter & range\\
            \hline
            \hline
            \texttt{Along track distance}          & {[}0.08, 3{]} NM    \\
            \texttt{Vertical path angle}    & [-2.2, -3.8]°  \\
            \texttt{Lateral path angle} & {[}- 4, 4{]} °     \\ 
            \texttt{Yaw}               & {[}-10,10{]} °   \\
            \texttt{Pitch}             & {[}-8,0{]} °     \\
            \texttt{Roll}              & {[}-10,10{]} °   \\ 
            \hline
        \end{tabular}
    }
    \captionof{table}{Parameters of the generic landing approach cone}
    \label{tab:odd_params}
\end{table}

These 6 parameters allow to define a generic landing approach cone (Definition~\ref{def:generic_approach_cone}) corresponding to a realistic aircraft trajectory during landing, as well as an envelope for the aircraft attitude that encompasses typical aircraft orientations during approaches on a runway.

\begin{definition}[Generic landing approach cone] A generic landing approach cone is the set of all pairs $\langle$positions, attitude$\rangle$ within the ranges of the 6 parameters of Table~\ref{tab:odd_params}.
\end{definition}\label{def:generic_approach_cone}

In addition to the approach cone specification, it is also relevant to define some \emph{operational scenarios} that describe some usual trajectories observed in the real world. 
Such scenarios can represent complex landing situations (e.g. crab and de-crab manoeuvres in the presence of wind)
or can be constructed by collecting real traffic observations \cite{opensky,olive2019quantitative}.
These scenarios will help in assessing the performance reached by of the ML constituent
and the safety of the complete system.

\begin{odd}[of VBL]
The VBL system must permit 
the landing as long as the aircraft is in the generic landing approach cone.
\end{odd}

\section{VBL ML constituent}
\label{sec-MLC}
The constraints of the system-level ODD expressed on the Operating Environment must be propagated to the components of the system. 
It is then mandatory to specify the ML constituent architecture
and its associated intended function.
The VBL constituent is expected to realise the following intended function: 
\begin{intended}[VBL intended function]
 \label{function}
  The \emph{intended function} is the pose estimation of the aircraft with respect to the airport runway when the aircraft flies within the generic landing approach cone. The pose is estimated from several sensors, including a camera positioned at the aircraft's nose and directly facing the runway during the landing. 
\end{intended}

\begin{figure}[h!bt]
    \centering
    \includegraphics[width=\linewidth]{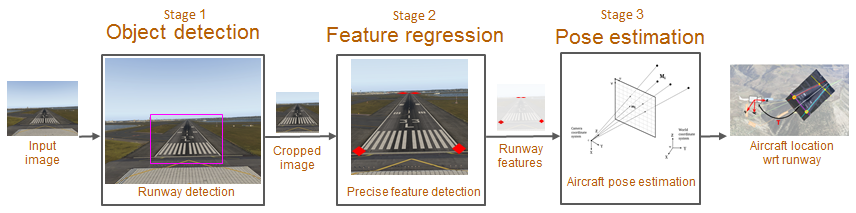}
    \caption{VBL constituent architecture with 3 stages}
    \label{fig:archi}
\end{figure}

The ML constituent architecture should fulfil the intended function. 
The one we propose is directly inspired from
the 3-stage architecture of Daedalaen AG \cite{faa_report} 
as shown in Figure \ref{fig:archi}. 
The first stage is based on an object detection step that is in charge of computing a bounding box around the detected runway. The image is then cropped around the bounding box and a second stage is in charge of computing the 4 corners of the runway. From this identification, the pose estimation of the aircraft is done with a non-ML approach by the last stage.

Subsequently, we solely focus on the first stage of the architecture that we call
the \emph{VBRD} (for Vision-Based Runway Detection).
This component comes with its own intended function and its associated performances.
The contribution of all MLC components intended functions and their associated performances
should ensure the MLC intended function and should 
fulfil the system requirements.
How to derive the requirements on each component and how to define the rules to combine the contributions of each component is out of scope of the paper.
These activities are non trivial and should benefit from a meticulous work.

The \emph{VBRD} is a pure ML component and 
is in charge of realising  an \emph{ML object detection task}.  
The family of object detection models covers a large range of applications and offers much more capacities than we need.
Indeed, such a model must be able to detect several types of objects (e.g. pedestrians, vehicles) and several objects appearing in a high resolution input image.
In our case, the model should identify one or more runways and
as a consequence, there is a unique class (runway). Moreover, we expect the output of the stage-1 to select the runway on which the aircraft must land.
An object detector, without any further support, will hardly be capable of selecting the target runway, in particular when there are parallel runways. 
To simplify the task, we decided to restrict the intended function
and thus the ODD
to have a unique runway on the image.
\begin{odd}[of VBL]
We restrict the operating conditions of the intended function \ref{function}.
\begin{enumerate}[topsep=1pt,itemsep=0pt, partopsep=0pt, leftmargin=*]
  \item The aircraft is landing on airports with a \emph{piano}; 
  \item There exists only one runway for which current position is considered within the approach cone\footnote{Another runway can still be visible, but the aircraft should not be in its approach cone};
  \item The runway is fully visible on the image (no occlusion).
\end{enumerate}
\end{odd}
Such restriction must be fed back to the system level for negotiation
and update on the system architecture / operating conditions.
In terms of object detection, the model belongs then to the category of
single class single object detection.
\begin{mytask}[Single object detection task]
 \label{task:object}
The detector must localize the object within the image 
 by providing \emph{bounding box} candidates surrounding said object.    
 An \emph{acceptable} bounding box should include the complete runway with a margin of x pixels on each side (top, bottom, left, right).
\end{mytask}

The VBRD  takes as input an image and outputs a bounding box around the runway it contains. Therefore, propagating the system-level ODD down to the components consists of characterising the appropriate constraints at the image level. 
At this stage, we can only define a very basic and un-exploitable image-level ODD
a first \emph{characterisation of the ML constituent restricted to the object detection task ODD}.
\begin{odd}[of task \ref{task:object}]
The ODD is the infinite set of all images that could be seen during a landing on an (extremely large) set of airport runways.
\end{odd}

\section{Image-level ODD}
\label{sec-image-level-odd}
To define properly an exploitable image-level ODD, we need to make more activities.

\subsection{ED-324/ARP6983 -- MLCODD Characterization and validation}
The MLCODD is defined from the operating environment identified at the system level to specify all foreseeable operating conditions under which the MLC is expected to work. 
Roughly speaking, we can see the MLCODD as a set of \emph{parameters} (or \emph{features}).
For instance, a \emph{parameter} could be the weather conditions selected
at system-level (e.g. the VBL should function correctly from -5$^\circ$C up to 40$^\circ$C in the presence of mist or in a perfect sunny day).
The parameter must then been 
translated as an image-level \emph{parameter}. For instance,
the weather conditions \emph{parameter} can be translated into several parameters such as contrast or brightness on the images. 
However, this translation into MLC inputs must be supported by some rationale and its impact on system-level parameters should also be examined. 
The MLCODD characterisation is instrumental in specifying the inputs of the data management process, i.e. the capture of all the requirements necessary to produce and verify the dataset. 

In any case, \emph{parameters} 
must be representable with a recognised nomenclature and understandable by a human. 
This entails that parameters must belong to well-typed elements (e.g. continuous parameters, set of nominal values for discrete/categorical parameters). 
After this preliminary identification of parameters, it may be that the number of image-level parameters is too high to be tractable. 
In that case, the MLCODD is refined by identifying potential interdependence between the parameters 
and applying a reduction strategy on the parameters to reduce the complexity and the dimension of the MLCODD.
In this paper, we consider geometric strategies and expert concepts identification to identify the parameters.

In addition, the MLCODD parameters are also characterised by some Data Quality Requirements (DQRs). 
Such a DQR can specify some ranges of values and a distribution where applicable. 
Regarding the example of the weather conditions \emph{parameter}, 
contrast or brightness parameters must be defined with a reachable range covering all the supported weather conditions and distribution among the range. 
Such a distribution can be conditioned on the airport (e.g. Toulouse and Montreal airport weather conditions distributions differ).
In this paper, we consider a subset of identified DQRs:
Source Suitability, Completeness, Representativeness 
and Accuracy. These properties will be detailed in Section~\ref{subsec:dqr_for_vbl}.

The ODD is not only limited to nominal situations. Indeed, the system must ensure safe behaviour in all foreseeable situations. As a result, the ODD must encompass more general cases. 
The ED-324/ARP6983 has defined its own taxonomy of data types for non-nominal data (outliers, edge/corner case, singularity, novelty) that should be considered with the appropriate stopping criteria.

\subsection{Approach to design an image-level ODD}
\label{subsec-img-odd}
In addition to the ED-324/ARP6983 considerations, it is worth looking at other works in the literature 
and other domain existing standards to help the designer in such a complex activity.
For instance, the ISO 34503 \cite{iso34503} encourages the designer to consider, in addition to parameters mentioned previously, 
\emph{elements} that correspond to main parts of an image. There are 2 categories of elements:
\begin{itemize}[topsep=2pt,itemsep=-1pt, partopsep=0pt, leftmargin=*]
    \item \emph{scenery elements} that refer to the spatially fixed elements of the operating environment relative to the aircraft;
    \item \emph{dynamic elements} that refers to moving elements (e.g. other aircraft).
\end{itemize}
To define the \emph{parameters} and the \emph{elements} of the VBRD,
we propose an
approach based on 3 activities, that is generic enough to be applied to other object detection ML constituent.

\subsubsection{\textbf{Geometry parameters}}
For now, the only usable constraints from the system level are the ones expressed with precise ranges on the geometry of the landing, represented by the definition of the \emph{Generic landing approach cone}
see section~\ref{def:generic_approach_cone}. 
The majority of the constraints on the image space will, therefore, be related to the position and attitude of the aircraft, but the focal length\footnote{The resolution of the image and the expected position of the runway will depend on this parameter combined with the distance to the runway.} of the camera will have to be taken into account as well. Thus, using the geometry of the landing, we can 
derive the possible positions of the runway on the image space. 
This activity should be supported by image processing methods. 
The book \cite{hartley2003multiple} recalls the basics of geometry for images and is a good basis for deriving some properties of the position/shape and other geometric considerations of the runway (or any other scenery element)
depending on the range of attitude of the aircraft. 
Among the transformations, we can cite the projection from the real world to the image-based coordinate system, which is done using two standard matrices \cite{szeliski2022computer}:
\begin{itemize}[topsep=2pt,itemsep=-1pt, partopsep=0pt, leftmargin=*]
    \item The Extrinsic matrix whose role is to get the coordinates of the corners in the camera-centered coordinate system.
    \item The Intrinsic matrix whose role is to project the 3D coordinates expressed in the camera-centered coordinate system into the 2D image.
\end{itemize}

\subsubsection{\textbf{Domain-specific concepts}}
Human making-decision process on an image relies on the identification of \emph{concepts}.
We propose the following partitioning of concepts, depending on their utility and relevance to the task of object detection:
\begin{itemize}[topsep=2pt,itemsep=-1pt, partopsep=0pt, leftmargin=*]
    \item \textbf{Primary concepts:} refer to elements (or landmarks) that are considered fundamental by a human for fulfilling the task. The absence of only one of the primary concepts would imply that the object considered is not an object of interest. For the detection of a runway, we can typically consider the shape of the runway (the typical 4-sided polygon), the clear demarcation with the external area, and the main markings (the target, the piano, the runway number)\footnote{This will be challenged when considering the distance at which the runway may be detected.}.
    \item \textbf{Secondary concepts:} refer to elements that may reinforce a decision, but the absence of which is not prohibitive in the identification of the object. For runway detection, we consider the surrounding elements such as the airport traffic control tower, secondary markings which are not always present (for instance the displaced thresholds), other surrounding runways, other aircraft in parking phase, taxiways parallel to the runway, etc...
    \item \textbf{Tertiary concepts:} The presence or the absence of the elements considered in this category should not have any impact on the detection of an object. We can identify here the colours of the areas surrounding the runway, due to vegetation or seasonal changes, as well as the environment around the airport itself, such as the presence of buildings, mountains or water bodies, etc...
\end{itemize}

A first analysis of these concepts highlighted that they could be different according to the distance to the runway. Indeed, the details on the image are not necessarily equivalent when the runway is a few kilometres away and when it is seen from a few hundred meters. In the first case, we may rely on secondary concepts like the overall airport and the traffic control tower, and for detecting the runway, we will rely in priority on the typical geometric shape of the runway and its visible markings (the target for instance). In the second case, we may consider the details of the markings, like the piano and the runway number, as well as parts of the secondary concepts like the markings on the surrounding taxiways.

We want to point out that these defined concepts could be used and extended in the validation phase of the Model. Indeed, if we consider a correct detection from an ML model, it is possible to use explainability methods such as~\cite{fel2023craft,ghorbani2019towards} to identify concepts which were used for this decision. These concepts can be categorised using the partition presented above as follows:
\begin{itemize}[topsep=2pt,itemsep=-1pt, partopsep=0pt, leftmargin=*]
    \item The concept belongs to the category of primary concept, and it should be added to the list of primary concepts already identified.
    \item The concept corresponds to a secondary one, which may or may not be used by humans. For instance, for detecting runways, we identified aircraft tyre marks as a valuable indication that is, consciously or not, used by humans when attempting the same task. However, since this concept is secondary, it is crucial that the model remains robust to its removal or its absence, as all runways are not heavily marked by tyres. Another example would be the difference in colours between the runway and its surroundings, or the possible texture of the runways, which can hardly be seen with the naked eye.
    \item The concept belongs to the tertiary ones, which means that it should not have been used for the decision-making process. In that case, it undoubtedly represents a bias of the model that should be eliminated, possibly by making changes to the training data, or by working on the retraining of the model.
\end{itemize}

For now, we only mentioned the correct detections of the ML model. Besides, in an ideal world, the list of primary concepts should be precise enough so that the absence of one concept would allow us to discriminate a true positive from a potential false positive. However, as mentioned earlier, this is not always true, typically due to the variability in the distance to the runway, which may lead to some runway features disappearing. 
Indeed, if we compare a highway segment to a runway from a relatively short distance, certain primary concepts such as geometry and shape, as well as the clear demarcation with the outer area are present, but the markings are very different. 
Nevertheless, at a certain distance, these markings are likely to be invisible, making it hardly possible to rely solely on these primary concepts to distinguish true positive detections from false positive ones. 
In that case, we can consider the problem in reverse: the false positive observed in the decision of a model could help us build a fourth list of concepts corresponding to well-identified biases leading to these potentially incorrect decisions. 

\textbf{Quaternary concepts:}
For objects identified as false positives by a model, we could include in the fourth list the set of elements that are primary concepts for this specific object (e.g. a highway) but not for our object or interest (e.g. a runway). For a highway, we could typically add the presence of cars, a central road divider, road markings, traffic signs, etc... This fourth list of concepts will, therefore, correspond to the elements which, if they are present, invalidate a detection.


\subsubsection{\textbf{DQR for VBL}}\label{subsec:dqr_for_vbl}
The guidelines \cite{ddsdeel}
 gives a set of recommendations to build and manipulate the datasets used to develop and/or validate machine learning models.

The Data source suitability \cite{wg114}
 ``\textit{refers to the appropriateness and relevance of a data source for a specific purpose or context, particularly in relation to its ability to provide data satisfying specified data quality attributes for a given task or analysis.}''

\begin{dqr}[Data source suitability for VBRD]
\label{dqr-suitability}
    This property is critical for the choice of generator of synthetic images which should be compared to the images captured by a camera in real conditions.
\end{dqr}

Completeness \cite{wg114}
is ``\textit{the extent to which a dataset covers, according to the specified criteria, the ODD for the intended application.}''

\begin{dqr}[Completeness]
\label{dqr-completeness}
    In our case, guarantees should be provided regarding the coverage of the operating conditions and the operational scenarios defined in system-level ODD.
\end{dqr}

Representativeness \cite{wg114} states that
``\textit{a dataset is representative if it covers the full ranges of the parameters that define the ODD and the distribution of each parameter matches the specified distribution.}''

\begin{dqr}[Representativeness]
\label{dqr-representativeness}
    This crucial property should motivate extensive testing of the data distribution regarding each parameter of the image-level ODD.
\end{dqr}



Accuracy/Correctness \cite{wg114}
"\textit{Measures the faithfulness to the real value and depends on data gathering/generation. It also measures the degree of ambiguity of the representation of the information.}".

\begin{dqr}[Accuracy/Correctness]
\label{dqr-correctness}
    In our case, this is highly related to the choice of labels and their precision for the task that must be fulfilled. 
\end{dqr}

These requirements are not guidelines to produce a dataset but must be kept in mind when designing it, ensuring a successful verification in later stages of the process.

\section{Design of the dataset}
\label{sec:dataset}
The objective is to design a dataset for the VBRD component compliant with image-level ODD defined by the activities of section \ref{subsec-img-odd}.

\subsection{ED-324/ARP6983 -- Data management and verification processes}
The objective of the \emph {data management process}  is to produce the dataset that matches the characterized ODD. The first activity of the data management process is to identify the sources of such data. Then data are collected, prepared and split into datasets in order to deliver trustworthy training, validation and test datasets which will be used to design, implement and integrate the ML inference model that meets the functional and operational requirements. 



Once the ODD is defined, it should be validated. That is the purpose of the \emph {ML data validation process} that is intended to provide assurance that the ODD and its DQRs are correct and complete with respect to the intended function supported by the MLC. 
The high quality of the datasets (and at least the test dataset for low critical applications) should be demonstrated. If this verification is not properly performed, then the trained model might exhibit unintended behaviour (e.g. make incorrect
decisions, fail to generalize to new or unseen situations) that could be detrimental to its intended use and/or the safety objectives that have been assessed at system level.
To this purpose, the ED-324/ARP6984 proposes several activities:

\begin{itemize}[topsep=2pt,itemsep=-1pt, partopsep=0pt, leftmargin=*]
    \item \emph{ODD/datasets bi-directional traceability} to guarantee that the complete ODD is covered and eliminate any undesirable data
    \item \emph{Data quality analysis} to demonstrate the datasets compliance to the DQRs.

\end{itemize}




 \subsection{Strategy to generate the dataset}\label{subsec:strategy_generation}
\begin{figure}[hbt] 
\centering 
    \resizebox{\linewidth}{!}{%
        \begin{tikzpicture}[thick,scale=0.5]
\usetikzlibrary{shapes,shapes.geometric,calc}
\usetikzlibrary[shadows]

\tikzstyle{block} = [draw,minimum height=2em, minimum width=1.5cm, inner sep=3pt];
\tikzstyle{txt} = [text centered, inner sep=0pt];
\tikzstyle{compliance} = [dashed, draw,  inner sep=3pt ];
\tikzstyle{database} = [cylinder, 
        shape border rotate=90, 
        draw,
        minimum height=1cm,
        minimum width=2cm,
        shape aspect=.25,];

\tikzset{
  multidocument/.style={
    shape=tape,
    draw,
    fill=white,
    tape bend top=none,
    double copy shadow},
  manual input/.style={
    shape=trapezium,
    draw,
    shape border rotate=90,
    trapezium left angle=90,
    trapezium right angle=80}}

\path (0,-2) node[database, fill= lightgray] (airportbase) {
\begin{tabular}{c} runway\\ database \end{tabular}
};

\path (airportbase)+(0,4) node[block, fill= lightgray, dashed] (confi) {
\begin{tabular}{c} Configuration\\ file \end{tabular}
};

\path (airportbase)+(6.5,2) node[block] (scengen) {
\begin{tabular}{c} Scenario\\generation \end{tabular}
};

\path (scengen)+(5.5,0) node[block] (ges) {
\begin{tabular}{c} Synthetic\\images\\environment \end{tabular}
};

\path (ges)+(5,0) node[block] (label) {
Labelling
};

\draw  [->]   (airportbase.east)  -| ($(airportbase.east)+(1,2)$) -- (scengen.west);

\draw  [->]   (confi.east)  -| ($(confi.east)+(1,-2)$) --(scengen.west);
\draw  [->]   (scengen.east)  -- (ges.west);
\draw  [->]   (ges.east)  -- (label.west);

\path (label)+ (8,0) node[database, fill= lightgray,minimum height=2.5cm,minimum width=5.5cm] (res) {
};
\path (res)+(0,2) node (text) {
Synthetic images
};

\draw  [->]   (label.east)  -- (res.west);

\path (res) + (-3,0) node [multidocument] (img) {image.jpg};

\path (img) + (5,0) node [multidocument] (lab) {\begin{tabular}{c} labels \\+ metadata\end{tabular}};

\end{tikzpicture}
    }
\caption{Generator pipeline} 
\label{fig-gen-pipeline} 
\end{figure}
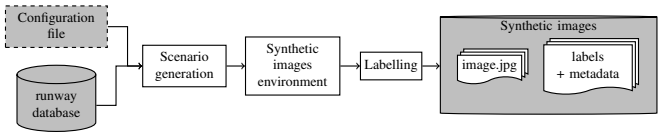

 \lard is composed of both synthetic and real footage images.
 Synthetic images were generated via a generator
generator pipeline presented in Figure \ref{fig-gen-pipeline}.
The two inputs (in gray) are the airport database and the configuration file to be filled by the user, setting which runway they want to generate images from and other parameters (e.g. number of images). Then, the first script (in white) generates a scenario file that can be provided as an input for the synthetic image environment (either \GES or \flightsim). 
This virtual globe tool can then generate the corresponding images, together with an information file  (here in json format). Finally, the last module of our generator associates the '\textit{labels}' to each image, in particular the scaled position of the four corners on the picture. The output in gray contains the images, the labels and the metadata.

Overall, this generator allows to produce an infinity of images with various camera angles and positions, where the annotation is automatically propagated, which drastically reduces the labelling cost.
In parallel, we have access to numerous Youtube channels from which we can extract images and label them.
\newline

This framework allows us to generate the dataset for our intended function. The question that we will have to tackle is how to generate sufficient representative data that fit the \emph{development of the data quality requirements}. 
As an example, the risk of having thousands of images per runway or more is the high similarity of resulting positions in the cone and the low independence between each image, which may lead to overfitting models, while collecting only a few dozen images per runway limits the possibility to encounter edge cases for each parameter and increases the need for manual annotation of runway corners to fulfil the high volume of data required. 
\subsection{Adequation between image-level ODD and DQRs}

The verification of the DQRs is a fundamental step to provide a first estimation of the quality of the proposed dataset. We detail in the following the quality analysis performed on the data and the results of the verification activities performed for each of the requirements defined in Section~\ref{subsec:dqr_for_vbl}.

\subsubsection{Source Suitability}

The task targeted by the ML component is the detection of runways on images during landing. However, the cost of labelling real images in a sufficient volume for ML training is prohibitive, which leads us to choose a tool for generating synthetic images instead. We selected Google Earth Studio, which supports trajectories of positions (defined within our landing approach cone) as input and allows us to produce a variety of high-quality images relatively close to reality. However this tool came with restrictions such as the absence of adverse weather simulation or realistic night images. As these constraints were not considered critical, we had to propagate them to the system-level ODD, producing a third refinement defined as follows:
\begin{odd}[of VBL]
We further restrict the operating conditions of the intended function~\ref{function}.
\begin{enumerate}[topsep=1pt,itemsep=0pt, partopsep=0pt, leftmargin=*]
  \item Optimal conditions: clear daylight and no adverse weather conditions (clouds, precipitations...).
\end{enumerate}
\end{odd}

\begin{figure}[hbt]
\centering
\begin{minipage}{0.49\linewidth}
      \includegraphics[width=\linewidth]{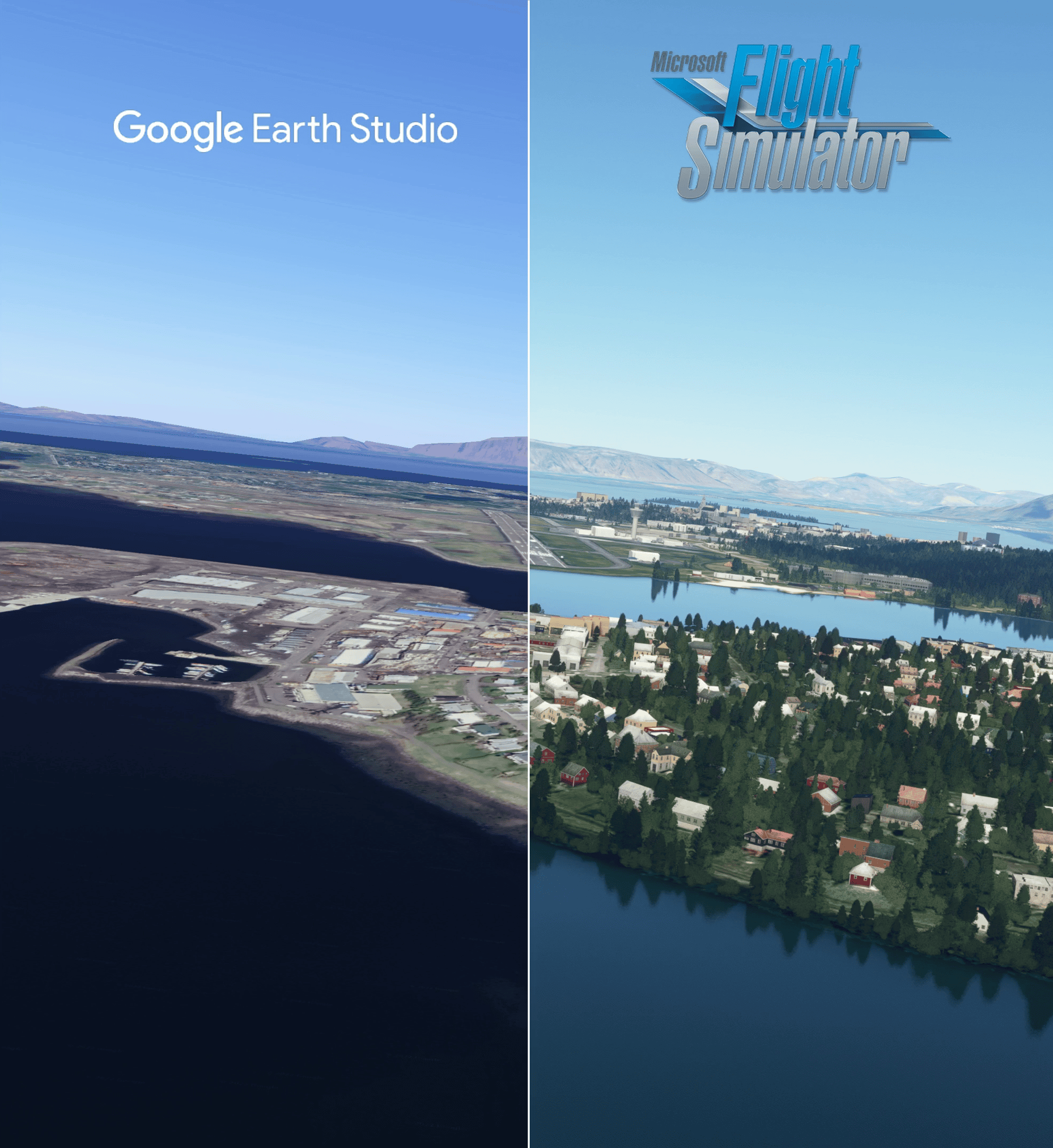}
    \caption{Comparison between Google Earth Studio and Flight Simulator}
    \label{fig:ges_vs_fsim}
\end{minipage}~
\begin{minipage}{0.49\linewidth}
   \includegraphics[width=\linewidth]{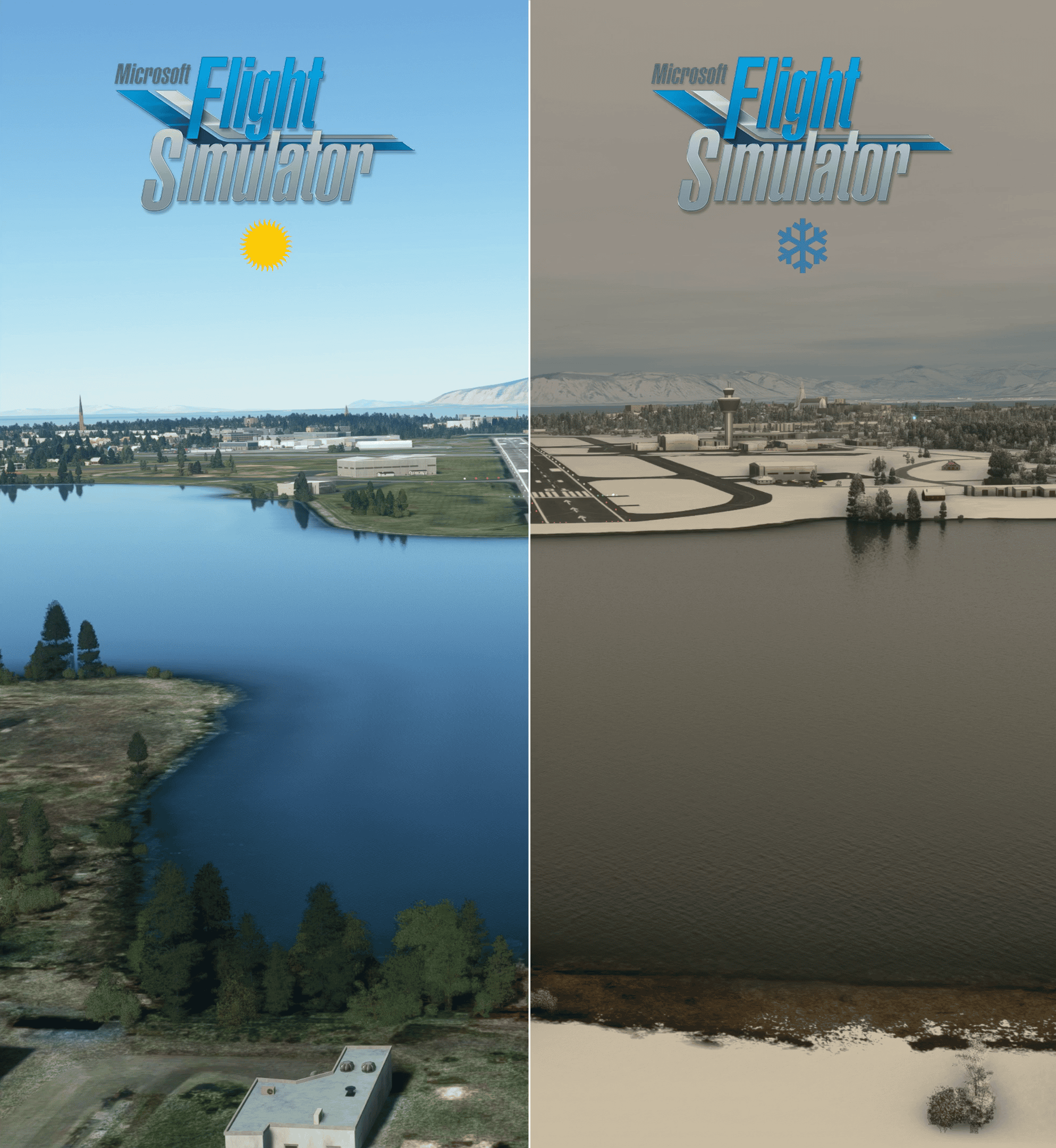}
    \caption{Illustration of weather and lighting variation on Flight Simulator }
    \label{fig:fsim_sun_snow}
\end{minipage}
\end{figure}

The quality analysis presented in this paper were performed on this refinement of the ODD. However, in a recent extension of our generator, we integrated the capability to generate images with both \GES and \flightsim, making this last restriction obsolete. Figure~\ref{fig:ges_vs_fsim} illustrates this comparison, while Figure~\ref{fig:fsim_sun_snow} shows some image variability supported in Flight Simulator.

To fulfil DQR \ref{dqr-suitability}, another complex question is the quality and the representativity of the synthetic images vs real images that will be observed at operation. There should be rationale and verification activity to accept synthetic images in the dataset. This is considered as future work.
\subsubsection{\textbf{Completeness}}
Considering the ODD of the VBRD, associated with the intended function~\ref{function}, an adequate dataset, i.e. that satisfies DQR \ref{dqr-completeness}, should not only cover a variety of airports all around the world but also span a wide range of positions inside the approach cone, to ensure a comprehensive coverage of all possible operational scenarios.



\begin{figure*}[t!]
    \centering
    \begin{subfigure}{0.38\textwidth}
        \includegraphics[width=\textwidth]{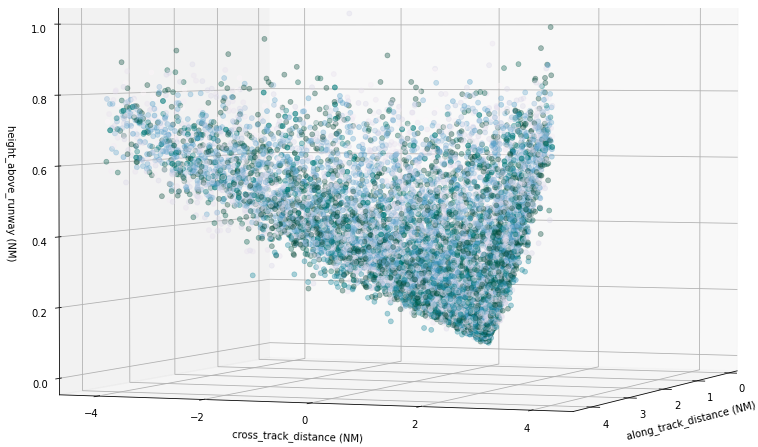}
        \caption{ \footnotesize Training set}        \label{fig:trajectories_of_landings_train}
    \end{subfigure}
    \hfill
    \begin{subfigure}{0.38\textwidth}
        \includegraphics[width=\textwidth]{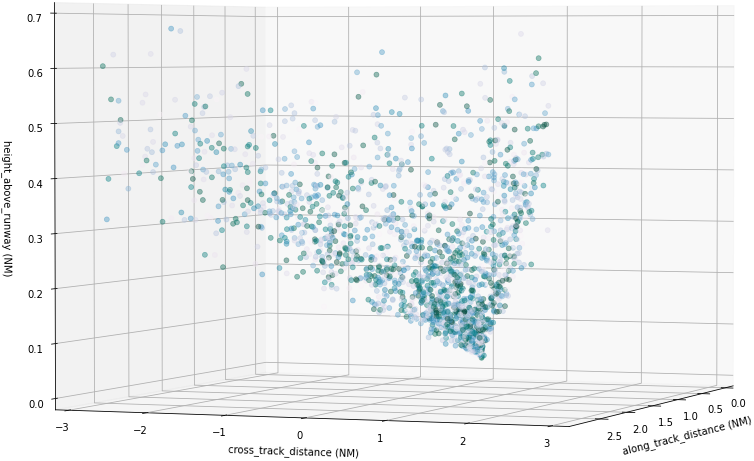}
        \caption{\footnotesize Synthetic test set}        \label{fig:trajectories_of_landings_test}
    \end{subfigure}
    \caption{3-dimensional visualisation of aircraft positions in the training set and the test set}
    \label{fig:trajectories_of_landings}
\end{figure*}

\noindent{\textbf{Coverage of landing scenarios:}}
Figure~\ref{fig:trajectories_of_landings} illustrates the distribution of aircraft positions in the training and the test set. 
In this figure, the z-axis corresponds to the \textit{along track distance}, but the other two axes are also distances (\textit{cross track distance} and \textit{height above runway}), computed from the angles provided in Table~\ref{tab:odd_params} (\textit{Lateral path angle} and \textit{Vertical path angle}). For the training set in Figure~\ref{fig:trajectories_of_landings_train}, the randomly sampled points span the whole approach cone corresponding to~\ref{def:generic_approach_cone}. 
Moreover, while the synthetic test set contains less data, it still covers a variety of positions in the cone, as illustrated in Figure~\ref{fig:trajectories_of_landings_test}.

\begin{figure}[hbt]
    \centering
    \includegraphics[width=.8\linewidth]{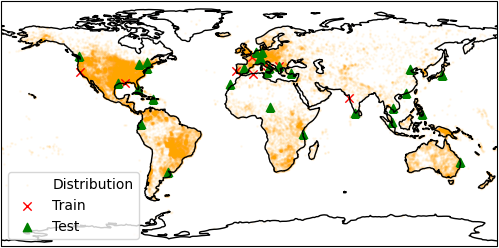}
    \caption{Distribution of airports used for the training set and the test set}
    \label{fig:airport_distribution}
\end{figure}

\noindent{\textbf{Airport distribution:}}
Figure~\ref{fig:airport_distribution} plots the distribution of airports from all around the world which were used to build the \lard dataset. Indeed, obtaining a great variety of images is a fundamental aspect for verifying the generalization capabilities of the models, and current distribution of airports presents the following benefits: 
first, it ensures a diversity of runway visuals, with different surface types\footnote{Asphalt and concrete are typically used for runway surfaces} and various runway length, width and markings, even if the runway standardization reduces the variability for this aspect. Second, it allows for a variety of surrounding terrain and landscapes such as grass, snow, dirt, but also city architectures, water bodies or mountainous reliefs.

\subsubsection{Representativeness}

For each parameter considered in the definition of the image-level ODD, we need to ensure that the distribution of the corresponding image-related features is thoroughly verified, as required by DQR \ref{dqr-representativeness}. 
We consider here that the test set is a faithful representation of our image-level ODD, and we compare the distribution of some of its features against the training set.

\footnotetext{Subset of the test set containing only images from real footage.}

\noindent{\textbf{Runway center positions:}} The plot of runway centers positions of Figure~\ref{fig:runway_centers_positions} shows an even distribution both for the training set and for the test set, located primarily around the center of the images. Nevertheless, a large area in the top and the bottom contain little to no points, which is the result of two main factors: 

\begin{figure}[ht]  
    \centering
    \includegraphics[width=.6\linewidth]{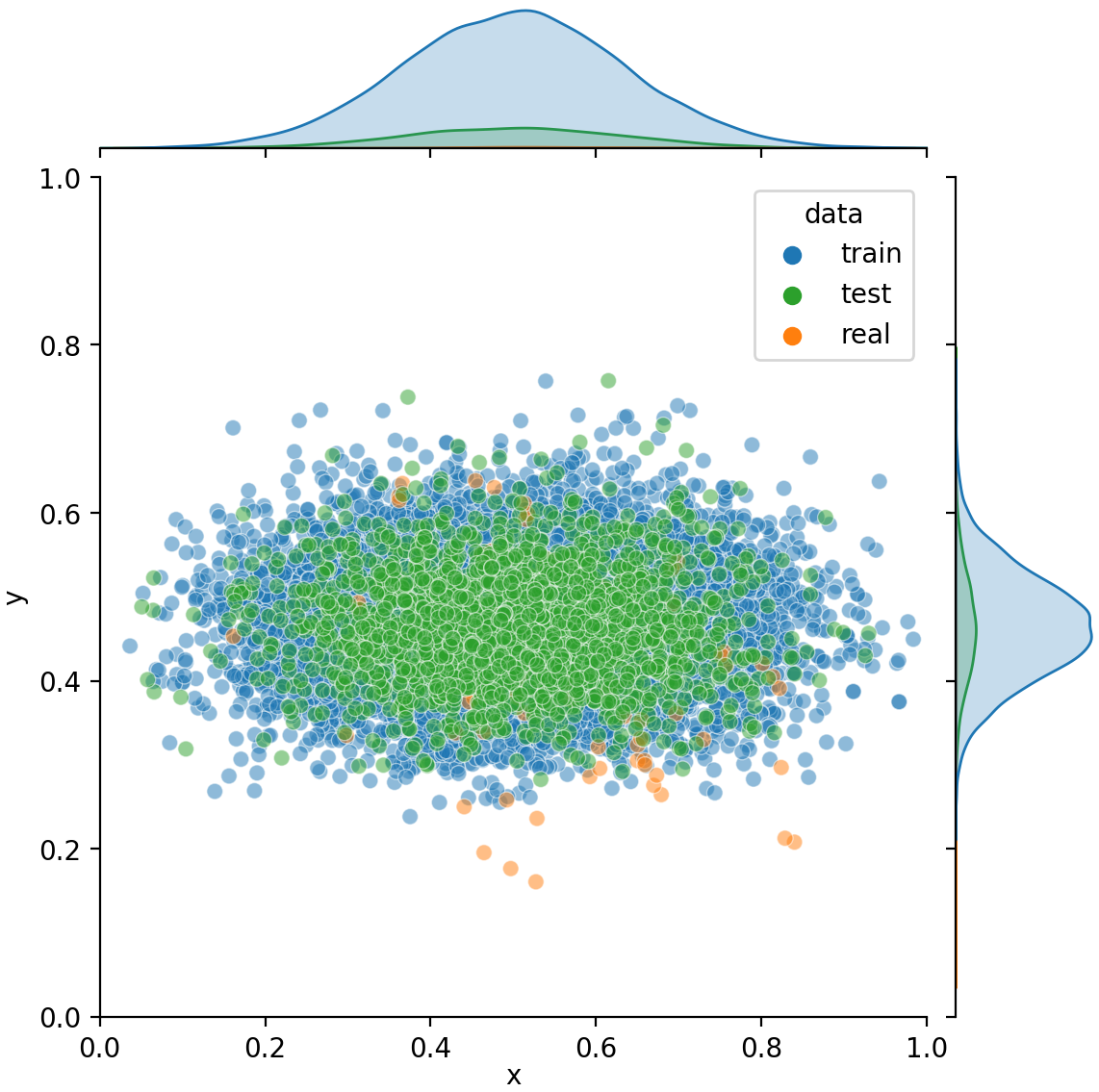}
    \caption[Normalized positions of runway centers in train and test and real subsets.]{Normalized positions of runway centers in train, test and real\protect\footnotemark~subsets.}
    \label{fig:runway_centers_positions}
\end{figure}

\begin{enumerate*}[label=(\roman*)]
    \item the presence of the watermark, which is expected to be removed from the images before usage by cropping 300 pixels from the top and the bottom of the pictures, and 
    \item the ranges of the \emph{pitch} parameter defined in the Table~\ref{tab:odd_params} which 
    prevent the runway to appear at the very top or bottom of the image.
\end{enumerate*}
Additionally, the real images of the test set appear to be slightly biased towards the bottom-right, which seems to result from the positions of the cameras in the cockpits.

\begin{figure}[ht]
    \centering
    \includegraphics[width=.8\linewidth]{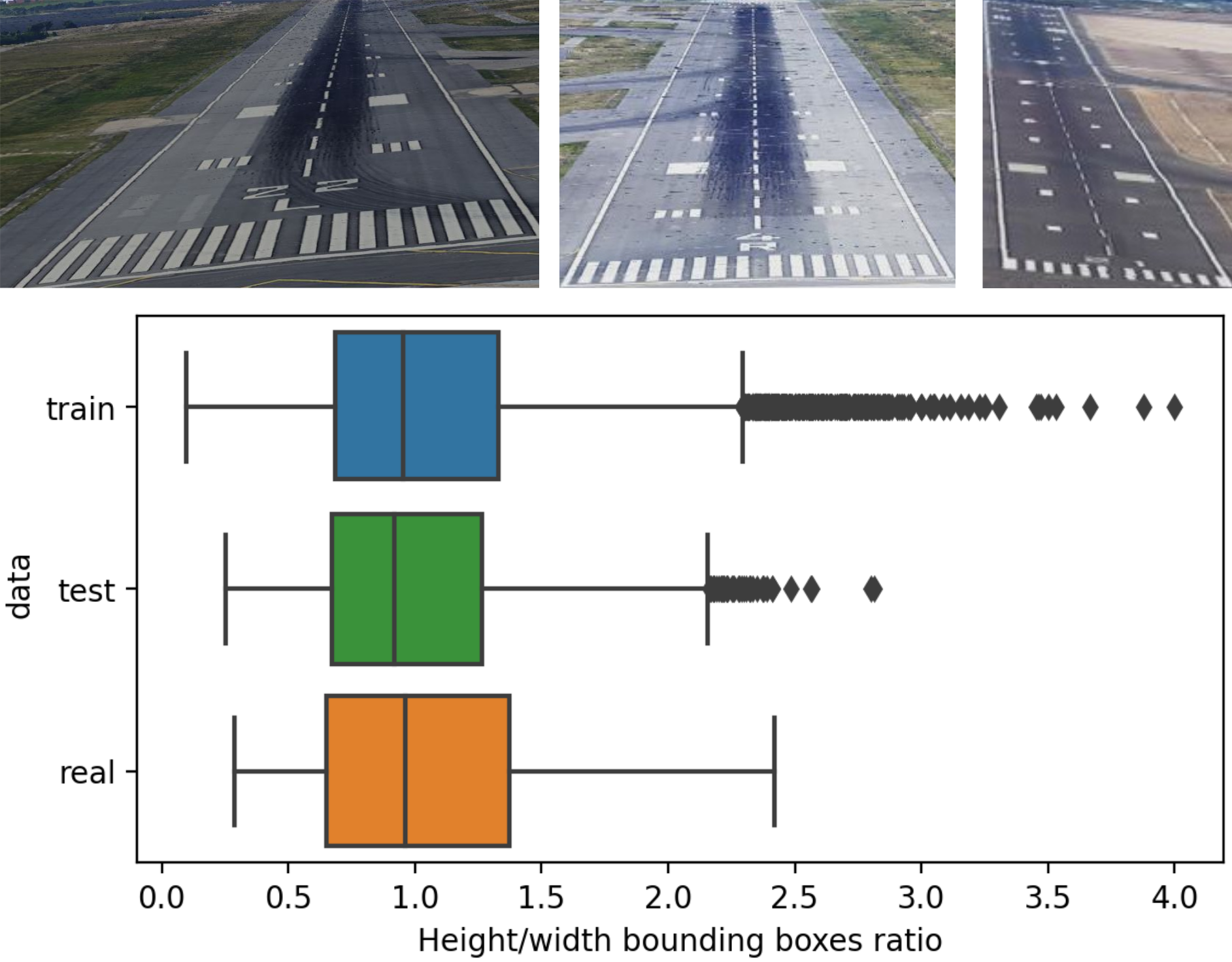}
    \caption{Top - \small{Illustration of different aspect ratios of the bounding boxes.} \normalsize{Bottom -} \small{Distributions of bounding boxes height over width ratios for the train, test and real subsets.}}
    \label{fig:bbox_ratio}
 \end{figure}

\noindent{\textbf{Bounding box fill ratios:}} The aspect ratio of the objects bounding boxes is a sensitive aspect for a detection task, as elongated objects in one or the other direction may not exhibit recognizable features. Figure~\ref{fig:bbox_ratio} illustrates the aspect ratio variability, and highlights how the majority of the bounding boxes in all three subsets have an aspect ratio between 0.5 and 1.5, indicating that most images are suitable for the targeted detection task.

\begin{figure}[hbt]
    \centering
    \begin{subfigure}{\linewidth}
        \centering
        \includegraphics[width=0.7\textwidth]{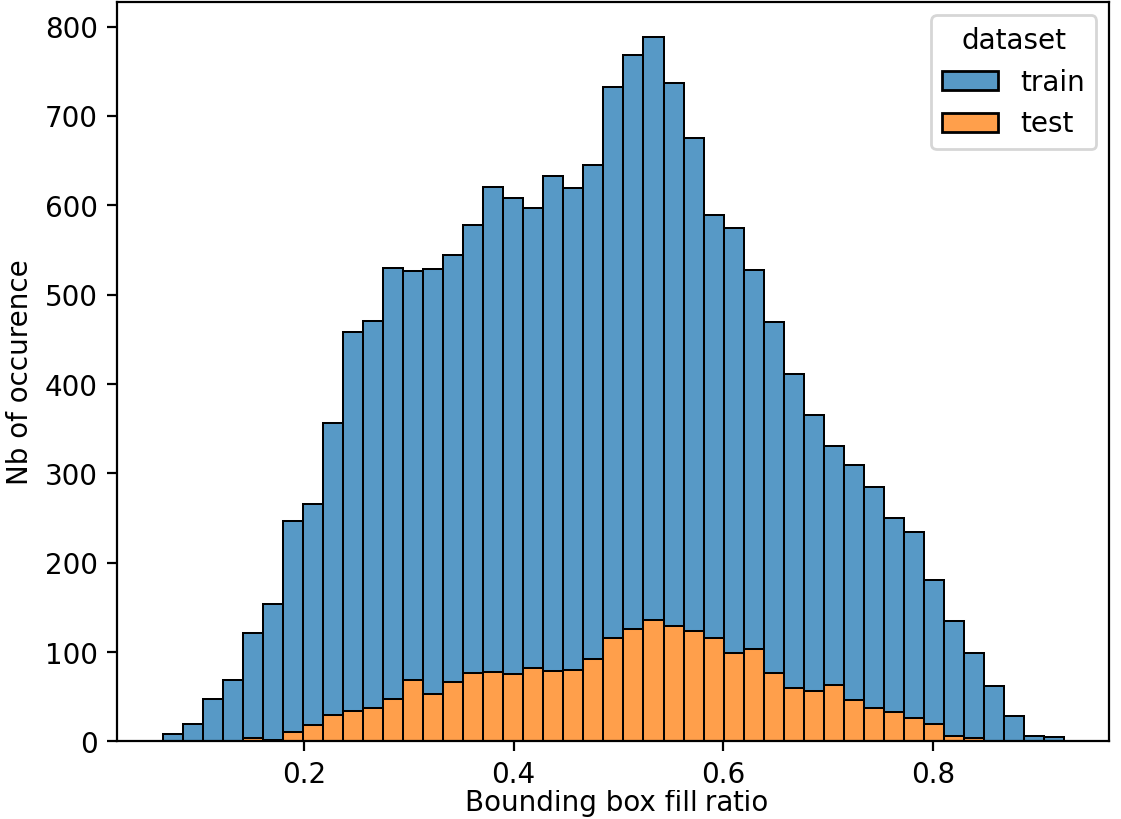}
        \caption{\footnotesize Distribution of bounding box fill ratios (percentage of the bounding box that correspond to pixels belonging to the runway itself)}
        \label{fig:bbox_filling_percentage}
    \end{subfigure}
    \begin{subfigure}{\linewidth}
        \centering
        \includegraphics[width=0.7\textwidth]{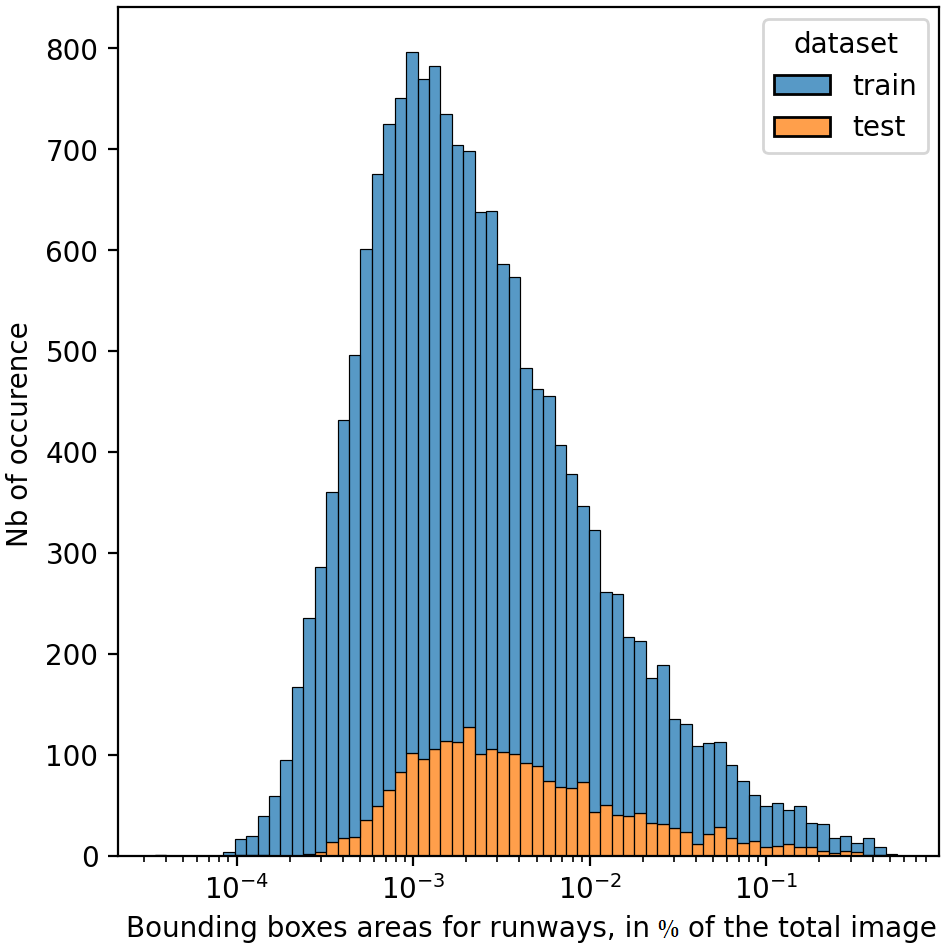}
        \caption{\footnotesize Distribution of bounding boxes areas (areas in logarithmic scale)}
        \label{fig:bbox_areas_percentage}
    \end{subfigure}
    \caption{Comparison of bounding box characteristics between training and test sets}
    \label{fig:histograms_bbox_comparison}
\end{figure}

The histograms of Figure~\ref{fig:histograms_bbox_comparison} illustrate the relationships between the runways, their bounding boxes and the global images. Figure~\ref{fig:bbox_filling_percentage} shows comparable distribution for the training and the test set, where most of the runways fill between 20$\%$ and 80$\%$ of their bounding boxes. This also indicates that bounding boxes should in general contain enough runways pixels for the detection task to be applicable and consistent. 
Additionally, Figure~\ref{fig:bbox_areas_percentage}, which illustrates how the areas of the bounding boxes cover the whole images, shows that the training set and the test set follow approximately the same distribution. 
This provides a certain level of guarantee that the bounding boxes will look similar between the training and the test set. 
Moreover, the figure shows that the vast majority of bounding boxes 
areas are over $25\times25$ pixels, which makes them large enough
 for a runway to be detected by humans.
On the other hand, the dataset contains only a few examples of bounding boxes with large size, which may bias the learning process when the aircraft is close to the runway 
and should be further investigated.

\noindent{\textbf{Slant distance:}} The synthetic images and the real images do not contain the same metadata. The distance between the aircraft and the runway is given for synthetic images as the \emph{slant distance}, however it is not available for real images, for which a value called \emph{time to landing} is provided instead. This value can be used as a proxy for the distance to the runway, considering that planes have comparable speed during landing phase.

\begin{figure}[ht]
    \centering
    \vspace{-10pt}
    \includegraphics[width=0.7\linewidth]{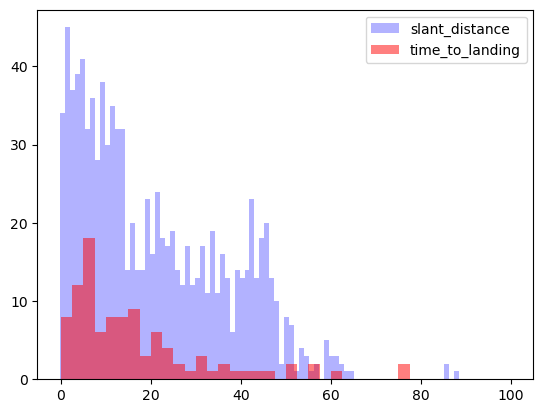}
    \caption{Comparison of distance estimation between real images and synthetic images in the test set}
    \label{fig:slant_distance_ttl}
\end{figure}

Figure~\ref{fig:slant_distance_ttl} shows how the distributions of \emph{slant distance} (for synthetic images) and  \emph{time to landing} (for real images) relate to each other\footnote{Only the shapes of the distributions should be compared as the \emph{slant distance} was re-scaled to fit the diagram}. It indicates that for both sources of data, the test set contains an important part of the images close to the runway while a non-negligible number of pictures were taken at longer distances from the runway, in a nearly evenly distributed manner, despite the limited number of real images.


\subsubsection{\textbf{Accuracy/Correctness}}

As specified by the ODD, the images of the dataset must always contain fully visible runways.
Any label associated to an image should allow to define the runway inside of it in an unambiguous way whether the data is synthetic or real footage, according to DQR \ref{dqr-correctness}. There are several approaches for delimiting a runway, the most usual being contours, ground marking, corners, horizon line or any other semantics specific to a runway.
We chose to encode the runway position by the pixel coordinates of its four corners in the image. 

As pointed out in~\cite{balduzzi2021neural}, representing the runway by its four corners poses some concerns such as instability in the presence of runway occlusion and sensitivity to the aircraft position estimation. In practice, none of the synthetic images (both in the training and the test sets) present situations of corner occlusions, but they may occur in real images, typically when the camera is placed inside the cockpit. 
However, these drawbacks are outweighed by the advantages of the corner representation, as this approach is easily applicable to any image, does not require camera-related information (typically camera angles), and is compatible with both image detection and image segmentation approaches, two of the most widely used approaches for locating and identifying objects in image. 

Finally, our labelling tool relies on the automatic projection of the runway corners onto the image, which ensures high precision of their position and directly satisfies the requirement for label accuracy.

\section{Conclusion}\label{sec:goals_scope}
Considering a well-defined vision-based landing task, we have presented a comprehensive approach for designing the system-level ODD of this intended function. We then focus on a specific ML component of the system and refine the ODD at the image level.  Using specific tooling based on Google Earth Studio, we illustrate the generation of a dataset designed to fulfil the targeted task. In the process, we establish a link with relevant standards relative to the introduction of ML-based systems in the industry. In particular, we focus on the definitions of Data Quality Requirements and show how they can be verified on the dataset defined for the intended function.

 The process leading to the definition of the ODD and the design of the dataset is an iterative process that also benefits from the model design.
Indeed, the verification of the model may exhibit some lacks in the dataset, e.g. insufficiency of images for certain airports or attitudes, that should be fed back to the data process management. We will investigate the whole process by pursuing the VBRD design following the ARP 6983 guidelines.

Moreover, our approach is complementary to~\cite{kaakai2023datacentric}, which defines the ODD by characterizing the distribution of samples it may contain (\textit{in} or \textit{out-of} ODD, edge-cases, corner cases...). We plan to leverage this formal definition to generate dedicated datasets for each of the parameters we identified in this paper, to investigate how the coverage of the ODD can be ensured.

Finally, the quality of a model solely trained on synthetic data must be carefully estimated. The addition of multiple data sources and more image variability should help us address this complex question in the future, but it would require a clear methodology that has yet to be written.



\section*{Acknowledgements}
{\small 
This work has benefited from the AI Interdisciplinary Institute ANITI,
funded by the “Investing for the Future – PIA3” program Grant agreement
ANR-19-P3IA-0004 and from the PHYDIAS 2 project funded by the French
government through the France Relance program, based on the funding from
and by the European Union through the NextGenerationEU program. The
authors gratefully acknowledge the support of the DEEL project\footnote{\url{https://www.deel.ai}}.
}

\bibliographystyle{unsrt}
\bibliography{bib}

\begin{thebibliography}{10}

\bibitem{gabreau:hal-WG}
Christophe Gabreau, Beatrice Pesquet-Popescu, and Fateh Kaakai.
\newblock {EUROCAE WG114 – SAE G34: a joint standardization initiative to
  support Artificial Intelligence revolution in aeronautics}, 2023.
\newblock Keynote of SafeAI --
  \url{https://safeai.webs.upv.es/wp-content/uploads/2021/02/AAAI_Keynote_202102_v1.0.pdf}.

\bibitem{gabreau:hal-03761946}
Christophe Gabreau, Adrien Gauffriau, Florence~De Grancey, Jean-Brice Ginestet,
  and Claire Pagetti.
\newblock {Toward the certification of safety-related systems using ML
  techniques: the ACAS-Xu experience}.
\newblock In {\em {11th European Congress on Embedded Real Time Software and
  Systems (ERTS 2022)}}, Toulouse, France, June 2022.

\bibitem{kaakai2023datacentric}
Fateh Kaakai, Shridhar~"Shreeder" Adibhatla, Ganesh Pai, and Emmanuelle
  Escorihuela.
\newblock Data-centric operational design domain characterization for machine
  learning-based aeronautical products, 2023.

\bibitem{Easaconcept}
{EASA}.
\newblock {Concept Paper: First usable guidance for Level 1 machine learning
  applications - Proposed Issue 01}, 2021.

\bibitem{Easaconcept2}
{EASA}.
\newblock {Concept paper: First usable guidance for Level 1 \& 2 machine
  learning applications - Proposed Issue 02}, 2023.

\bibitem{arp4754}
{SAE/EUROCAE}.
\newblock {Aerospace {R}ecommended {P}ractices {A}{R}{P}4754a/ED-79A-
  Development of Civil Aircraft and Systems}, 2010.

\bibitem{do178c}
{RTCA/EUROCAE}.
\newblock D{O}-178{C}/{E}{D}-12{C} - {S}oftware {C}onsiderations in {A}irborne
  {S}ystems and {E}quipment {C}ertification, 2011.

\bibitem{do254}
{RTCA/EUROCAE}.
\newblock {DO-254}/{E}{D}-80 - {D}esign {A}ssurance {G}uidance {F}or {A}irborne
  {E}lectronic {H}ardware, 2000.

\bibitem{ducoffe:hal-04056760}
M{\'e}lanie Ducoffe, Maxime Carrere, L{\'e}o F{\'e}liers, Adrien Gauffriau,
  Vincent Mussot, Claire Pagetti, and Thierry Sammour.
\newblock {LARD - Landing Approach Runway Detection - Dataset for Vision Based
  Landing}.
\newblock working paper or preprint, April 2023.

\bibitem{j3016}
{SAE}.
\newblock {J3016 Levels of Automated Driving}, 2019.

\bibitem{sotif}
{ISO}.
\newblock {ISO 21448:2022. Road vehicles — Safety of the intended
  functionality}, 2022.

\bibitem{iso34503}
{ISO}.
\newblock {ISO 34503:2023. Road Vehicles — Test scenarios for automated
  driving systems — Specification for operational design domain}, 2023.

\bibitem{Koopman2019HowMO}
Philip Koopman and Frank Fratrik.
\newblock How many operational design domains, objects, and events?
\newblock In {\em SafeAI@AAAI}, 2019.

\bibitem{runway_marking}
FAA.
\newblock {Airport Marking Aids and Signs}.
\newblock
  \url{https://www.faa.gov/air_traffic/publications/atpubs/aim_html/chap2_section_3.html},
  2022.

\bibitem{opensky}
Matthias Schäfer, Xavier Olive, Martin Strohmeier, Matthew Smith, Ivan
  Martinovic, and Vincent Lenders.
\newblock Opensky report 2019: Analysing tcas in the real world using big data.
\newblock In {\em 2019 IEEE/AIAA 38th Digital Avionics Systems Conference
  (DASC)}, 2019.

\bibitem{olive2019quantitative}
Xavier Olive and Pierre Bieber.
\newblock Quantitative assessments of runway excursion precursors using mode s
  data.
\newblock {\em arXiv preprint arXiv:1903.11964}, 2019.

\bibitem{faa_report}
{Daedalaen AG}.
\newblock {Neural Network Based Runway Landing Guidance for General Aviation
  Autoland}, 2022.

\bibitem{hartley2003multiple}
Richard Hartley and Andrew Zisserman.
\newblock {\em Multiple view geometry in computer vision}.
\newblock Cambridge university press, 2003.

\bibitem{szeliski2022computer}
Richard Szeliski.
\newblock {\em Computer vision: algorithms and applications}.
\newblock Springer Nature, 2022.

\bibitem{fel2023craft}
Thomas Fel, Agustin Picard, Louis Bethune, Thibaut Boissin, David Vigouroux,
  Julien Colin, R{\'e}mi Cad{\`e}ne, and Thomas Serre.
\newblock Craft: Concept recursive activation factorization for explainability.
\newblock In {\em Proceedings of the IEEE/CVF Conference on Computer Vision and
  Pattern Recognition}, pages 2711--2721, 2023.

\bibitem{ghorbani2019towards}
Amirata Ghorbani, James Wexler, James~Y Zou, and Been Kim.
\newblock Towards automatic concept-based explanations.
\newblock {\em Advances in neural information processing systems}, 32, 2019.

\bibitem{ddsdeel}
Cyril Cappi, Camille Chapdelaine, Laurent Gardes, Eric Jenn, Baptiste
  Lef{\`{e}}vre, Sylvaine Picard, and Thomas Soumarmon.
\newblock Dataset definition standard {(DDS)}.
\newblock 2021.

\bibitem{wg114}
{EUROCAE WG-114/SAE joint group}.
\newblock {ED 327/ARP 6983 -- Certification/approval of aeronautical systems
  based on {A}{I}}, 2023.
\newblock on going standardization.

\bibitem{balduzzi2021neural}
Giovanni Balduzzi, Martino Ferrari~Bravo, Anna Chernova, Calin Cruceru, Luuk
  van Dijk, Peter de~Lange, Juan Jerez, Nathana{\"e}l Koehler, Mathias Koerner,
  Corentin Perret-Gentil, et~al.
\newblock Neural network based runway landing guidance for general aviation
  autoland.
\newblock Technical report, United States. Department of Transportation.
  Federal Aviation Administration~…, 2021.

\end{thebibliography}

\end{document}